\documentclass[letterpaper, 10 pt, conference]{ieeeconf}  

\IEEEoverridecommandlockouts                              

\makeatletter
\newcommand*{\rom}[1]{\expandafter\@slowromancap\romannumeral #1@}
\makeatother

\usepackage[bindingoffset=0.2in,%
            left=0.75in,right=0.75in,top=0.75in,bottom=0.75in,%
            footskip=.25in]{geometry}
\usepackage{graphics} 
\usepackage{epsfig} 
\usepackage{epstopdf}
\graphicspath{ {figs} }
\usepackage{graphicx}
\usepackage{caption}
\usepackage{xcolor}
\usepackage{subcaption}
\usepackage{lipsum}
\usepackage{float}
\usepackage{todonotes}
\usepackage{color,soul}
\usepackage{hyperref}
\usepackage{amsmath} 

\DeclareMathOperator*{\argmax}{\arg\!\max}
\usepackage[scientific-notation=true]{siunitx}
\usepackage{amssymb}  
\usepackage{algorithm2e}
\usepackage{array}
\newcolumntype{x}[1]{>{\centering\arraybackslash\hspace{0pt}}p{#1}}

\newcommand{\bx}{\mathbf{x}}
\newcommand{\ba}{\mathbf{a}}
\newcommand{\bs}{\mathbf{s}}
\newcommand{\bI}{\mathbf{I}}
\newcommand{\cS}{\mathcal{S}}
\newcommand{\cX}{\mathcal{X}}
\newcommand{\cA}{\mathcal{A}}
\newcommand{\cB}{\mathcal{B}}
\newcommand{\cD}{\mathcal{D}}

\newcommand{\etal}{{\em et al.}}

\definecolor{dblue}{rgb}{0,0,0.7}

\title{\LARGE \bf
  Deep Reinforcement Learning with Successor Features \\ for Navigation across Similar Environments
}

\author{Jingwei Zhang \and Jost Tobias Springenberg \and Joschka Boedecker \and Wolfram Burgard
\thanks{All authors are with the University of Freiburg, Institute of Computer Science, 79110 Freiburg, Germany.
  This work was partly funded through the State Graduate Funding Program of
  Baden-W{\"u}rttemberg and by the DFG grant SPP-1597.}
\thanks{{\tt\scriptsize \{zhang,springj,jboedeck,burgard\}@cs.uni-freiburg.de}}%
}

\begin{document}

\maketitle
\thispagestyle{empty}
\pagestyle{empty}

\begin{abstract} In this paper we consider the problem of robot
  navigation in simple maze-like environments where the robot has to
  rely on its onboard sensors to perform the navigation task.
  In particular, we are interested in solutions to this problem
  that do not require localization, mapping or planning.
  Additionally, we
  require that our solution can quickly adapt to new situations (e.g.,
  changing navigation goals and environments).  To meet these
  criteria we frame this problem as a sequence of related reinforcement learning
  tasks. We propose a successor-feature-based
  deep reinforcement learning
  algorithm that can learn to transfer knowledge from previously
  mastered navigation tasks to new problem instances. Our algorithm
  substantially decreases the required learning time after the
  first task instance has been solved, which makes it easily adaptable
  to changing environments. We validate our method in both simulated and real robot experiments
  with a Robotino
  and compare it to a set of
  baseline methods including classical planning-based navigation.
\end{abstract}

\section{Introduction}
\label{sec:introduction}

Autonomous navigation is one of the core problems in mobile robotics. It can roughly be characterized as the ability of a robot to get from its current position to a designated goal location solely based on the input it receives from its on-board sensors. A popular approach to this problem relies on the successful combination of a series of different algorithms for the problems of simultaneous localization and mapping (SLAM), localization in a given map as well as path planning and control, all of which often depend on additional information given to the agent. Although individually the problems of SLAM, localization, path planning and control are well understood~\cite{Thrun05,Lav06,Lat91}, and a lot of progress has been made on learning control \cite{kober2013reinforcement}, they have mainly been treated as separable problems within robotics and some often require human assistance during setup-time. For example, the majority of SLAM solutions are implemented as passive procedures relying on special exploration strategies or a human controlling the robot for sensory data acquisition. In addition, they typically require an expert to check as to whether the obtained map is accurate enough for path planning and localization.

Our goal in this paper is to make first steps towards a solution for navigation tasks without explicit localization, mapping and path planning procedures. To achieve this we adopt a reinforcement learning (RL) perspective, building on recent successes of deep RL algorithms for solving challenging control tasks \cite{mnih2015human,lillicrap2016,kulkarni2016deep,lfda-e2e-15,icml2015_schulman15}.  For such an RL algorithm to be useful for robot navigation we desire that it can quickly adapt to new situations (e.g., changing navigation goals and environments) while still preserving the solutions to earlier problems: a prerequisite that is not fulfilled by current state-of-the-art RL-based methods. To achieve this, we employ \textit{successor representation} learning, which has recently also been combined with deep nets \cite{barreto2016successor,kulkarni2016deep}. As we show in this paper, this formulation can be extended to handle sequential task transfer naturally, with minimal additional computational costs; its ability of retaining a compact representation of the Q functions of all encountered tasks enables it to cope with the limited memory and processing capabilities on robotic platforms.

\begin{figure}[t]
    \begin{subfigure}{0.48\textwidth}
    	\centering
        \includegraphics[height=1.64in]{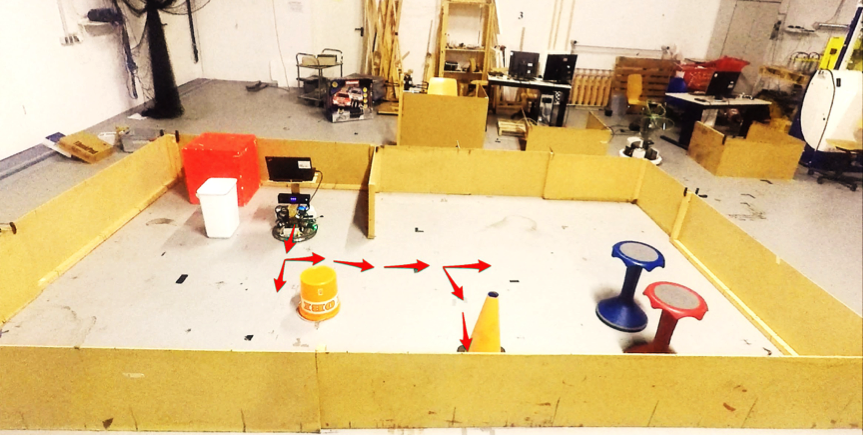}
    \end{subfigure}
    \vspace{0.1cm}
    \hrule
    \vspace{0.1cm}
    \begin{subfigure}{0.48\textwidth}
        \includegraphics[height=0.38in]{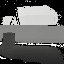}
        \includegraphics[height=0.38in]{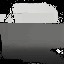}
        \includegraphics[height=0.38in]{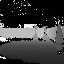}
        \includegraphics[height=0.38in]{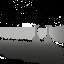}
        \includegraphics[height=0.38in]{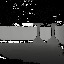}
        \includegraphics[height=0.38in]{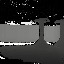}
        \includegraphics[height=0.38in]{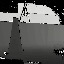}
        \includegraphics[height=0.38in]{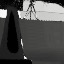}
    \end{subfigure}
    \caption{Exemplary maze-like environment considered in this paper (\textit{Map6}) and the optimal path from a randomly chosen start position to the goal (orange traffic cone) taken by the Robotino robot (top) together with the sensory input captured by the robot's on-board kinect sensor (bottom).}
    \label{fig:title-real}
\end{figure}

We validate our approach and its fast transfer learning capabilities in both simulated and real world experiments, on both visual and depth inputs, where the agent must navigate different maze-like environments. We compare it to several baselines such as a conventional planner (assuming perfect localization), a supervised imitation learner (assuming perfect localization during training only), and transfer with DQN. In addition, we validate that deep convolutional neural networks (CNNs) can be used to imitate conventional planners in our considered domain.

\section{Relations To Existing Work}
\label{sec:relatedWorks}
Our work is related to an increasing amount of literature on deep reinforcement learning. We here highlight the most apparent connections to recent trends with a focus on value based RL (which we use as a basis). A more detailed review of the concepts we built upon is then given in Sec.~\ref{sec:background}.

As mentioned, a growing amount of success has been reported for value-based RL in combination with deep neural networks. This idea was arguably popularized by the Deep Q-networks (DQN)~\cite{mnih2015human} approach followed by a large body of work deriving extended variants (e.g., recent adaptations to continuous control ~\cite{lillicrap2016,icml2015_schulman15} and improvements stabilizing its performance ~\cite{Hasselt2016,Wang2016,Schaul2015}).

While the DQN inspired RL algorithms were shown to be surprisingly effective, they also come with some caveats that complicate transfer to novel tasks (one of the key attributes we are interested in).
More precisely, although a neural network trained using Q-learning on a specific task is expected to learn features that are informative about both: i) the dynamics induced by the policy of the agent in a given environment (we refer to this as the \textit{policy dynamics} in the following text), and ii) the association of rewards to states; these two sources of information cannot be assumed to be clearly separated within the network.
As a consequence, while fine-tuning a learned Q-network on a related task might be possible, it is not immediately clear how the aforementioned learned knowledge could be transferred in a way that keeps the policy on the original task intact. One attempt at clearly separating reward attribution for different tasks while learning a shared representation is the idea of learning a general (or universal) value function~\cite{SuttonHorde11} over many (sub)-tasks that has recently also been combined with DQN-type methods~\cite{SchaulUVFA}. Our method can be interpreted as a special parametrization of a general value function architecture that facilitates fast task transfer.

Task transfer is one of the long standing problems in RL. Historically, most existing work in this direction relied on simple task models and explicitly known relations between tasks or known dynamics~\cite{ring1995,Taylor2011,WilsonFRT07}. More recently, there have been several attempts at combining task transfer with Deep RL~\cite{Parisotto2015,Rusu2015,rusu2016progressive,tai2016towards,Zhu16,tai2016deep}.
E.g., Parisotto \etal~\cite{Parisotto2015} and Rusu \etal~\cite{Rusu2015} performed multi-task learning (transferring useful features between different ATARI games) by fine-tuning a DQN network (trained on a single ATARI game) on multiple ``related'' games. 
More directly related to our work, Rusu \etal~\cite{rusu2016progressive} developed the \textit{Progressive Networks} approach which trains an RL agent to progressively solve a set of tasks, allowing it to re-use the feature representation learned on tasks it has already mastered. Their derivation has the advantage that performance on all considered tasks is preserved but requires an ever growing set of learned representations.

In contrast to this, our approach for task transfer aims to more directly tie the learned representations between tasks. To achieve this, we build on the idea of \textit{successor representation} learning for RL first proposed by Dayan~\cite{dayan1993improving}, and recently combined with deep neural networks ~\cite{barreto2016successor,kulkarni2016deep}. This line of work makes the observation that Q-learning can be partitioned into two sub-tasks: 1) learning features from which the reward can be predicted reliably and 2) estimating how these features evolve over time. While it was previously noted how such a partitioning can be exploited to speed up learning for cases where the reward changes scale or meaning~\cite{barreto2016successor,kulkarni2016deep}, we here show how this view can be extended to allow general -- fast -- transfer across tasks, including changes to the environment, the reward function and thus also the optimal policy.



We also note that the objective we use for learning descriptive features involves training a deep auto-encoder. Learning state representations for RL via auto-encoders has previously been considered several times in the literature~\cite{Riedmiller12,Jonschkowski2014,Finn2015}. Among these, utilizing the priors on learned representation for robotics from Jonschkowski \etal~\cite{Jonschkowski2014} could potentially further improve our model.


\section{Background}
\label{sec:background}
In this section we will first review the concepts of reinforcement learning upon which we build our approach.

\subsection{Reinforcement learning}
We formalize the navigation task as a \textit{Markov Decision Process} (MDP). In an MDP an agent interacts with the environment through a sequence of observations, actions and reward signals. In each time-step $t \in \lbrack0, T \rbrack$ of the decision process the agent first receives an observation from the environment $\bx_t \in \cX$ (in our case an image of its surrounding).
Together with a history of recent observations $\lbrace \bx_{t-H}, \dots \bx_{t-1} \rbrace$ -- with history length $H$ --, $\bx_t$ informs the agent about the true state of the environment $\bs_t \in \cS$. In the following always define $\bs_t$ as $\bs_t = \lbrace \bx_{t-H}, \dots \bx_{t-1}, \bx_t \rbrace$.
The agent then selects an action $\ba_t \in \cA$ according to a policy $\ba_t = \pi(\bs_t)$
\footnote{We restrict the following presentation to deterministic policies with discrete actions to simplify notation. A generalization can easily be obtained.}
and transits to the next state $\bs_{t+1}$ following the dynamics of the environment: $p(\bs_{t+1}|\bs_{t}, \ba_t)$, receiving reward $R(\bs_t) \in \mathbb{R}$ and obtaining a new observation $\bx_{t+1}$.
The agent's goal is to maximize the cumulative expected future reward (with discount factor $\gamma$). This quantity uniquely assigns an expected value to each state-action pair. The action-value function (referred to as the Q-value function) of executing action $\ba$ in state $\bs$ under a policy $\pi$ thus can be defined as:
\begin{align}
    Q(\bs, \ba; \pi) = \mathbb{E}\left[\sum_{t=0}^{\infty} \gamma^t R(\bs_t) \Big| \bs_0=\bs, \ba_0=\ba, \pi \right],
    \label{eq:q}
\end{align}
where the expectation is taken over the policy dynamics: the transition dynamics under policy $\pi$.
Importantly, the Q-function can be computed using the Bellman equation
\begin{equation}
Q(\bs_{t}, \ba_{t}; \pi) = R(\bs_t) + \gamma \mathbb{E}\left[ Q(\bs_{t+1}, \ba_{t+1}; \pi) \right],
\end{equation}
which allows for recursive estimation procedures such as Q-learning and SARSA~\cite{Sutton1998}. Furthermore, assuming the Q-function for a given policy is known, we can find an improved policy $\hat{\pi}$ by greedily choosing $\ba_t$ in each state: $\hat{\pi}(\bs_t) = \argmax_{\ba_t} Q(\bs_t, \ba_t; \pi)$.

When combined with powerful function approximators such as deep neural
networks
these principles form the
basis of many recent successes in RL for control.

%

\subsection{Successor feature reinforcement learning}
\label{sec:background_sf}
While direct learning of the Q-value function from Eq. \eqref{eq:q} with function approximation is possible, it results in a black-box approximator which makes knowledge transfer between tasks challenging (we refer to Sec.~\ref{sec:relatedWorks} for a discussion). We will thus base our algorithm upon a re-formulation of the RL problem first introduced by~\cite{dayan1993improving} called \textit{successor representation} learning which has recently also been combined with deep neural networks~\cite{barreto2016successor,kulkarni2016deep}, that we will first review here and then extend to naturally handle task transfer.

To start, we assume that the reward function can be approximately represented as a linear combination of learned features $\phi(\bs; \theta_{\phi})$ (in our case features extracted from a neural network) with parameters $\theta_\phi$ and a reward weight vector $\omega$ as $R(\bs) \approx \phi(\bs; \theta_{\phi})^T \omega$. Using this assumption we can rewrite Eq. \eqref{eq:q} as
\begin{align}
    Q(\bs, \ba; \pi)
  &\approx
    \mathbb{E}\left[ \sum_{t=0}^{\infty} \gamma^t \phi(\bs_{t}; \theta_\phi) \cdot \omega \Big| \bs_0=\bs, \ba_0=\ba, \pi \right]
\notag\\&=
    \mathbb{E}\left[ \sum_{t=0}^{\infty} \gamma^t \phi(\bs_{t}; \theta_\phi) \Big| \bs_0=\bs, \ba_0=\ba, \pi \right] \cdot \omega
\notag\\&=
    \psi^{\pi}(\bs, \ba)^T \omega,
\label{eq:srqfac}
\end{align}
where, in line with~\cite{barreto2016successor} we refer to $\psi^{\pi}(\bs, \ba) = \mathbb{E}\left[ \sum_{t=0}^{\infty} \gamma^t \phi(\bs_{t}  ; \theta_\phi) | \bs_0=\bs, \ba_0=\ba, \pi \right]$ as the \textit{successor features}. Consequently we will refer to the whole reinforcement learning algorithm as successor feature reinforcement learning (SF-RL).
As a special case we will assume that the features $\phi(\bs; \theta_{\phi})$ themselves are representative of the state $\bs$ (i.e., we can reconstruct the state $\bs$ from $\phi(\bs; \theta_{\phi})$ alone) which allows us to explicitly turn $\psi(\cdot)$ into a function $\psi^{\pi}(\phi(\bs_t; \theta_\phi), \ba_t)$. In the following we use the short-hand $\phi_\bs = \phi(\bs; \theta_\phi)$ -- omitting the dependency on the parameters $\theta_\phi$ -- and write $\psi^{\pi}(\phi_{\bs_t}, \ba_t)$  to avoid cluttering notation.

 Interestingly, these successor features can again be computed via a Bellman equation in which the reward function is replaced with $\phi_{\bs_t}$; that is we have:
\begin{equation}
\psi^{\pi}(\phi_{\bs_t}, \ba_t) =  \phi_{\bs_t} + \gamma \mathbb{E} \left[ \psi^{\pi}\left(\phi_{\bs_{t+1}}, \ba_{t+1} \right) \right].
\end{equation}
And we can thus learn approximate successor features using a deep Q-learning like procedure~\cite{barreto2016successor,kulkarni2016deep}. Effectively, this re-formulation separates the learning of the Q-function into two problems: 1) estimating the expectation of descriptive features under the current policy dynamics and 2) estimating the reward obtainable in a given state.

To show how learning with successor feature RL works let us consider the case where we are only interested in recovering the Q-function of the optimal policy $\pi^*$. In this case we can simultaneously learn the parameters $\theta_\phi$ of the feature mapping $\phi_\bs$ (a convolutional neural network), the reward weights $\omega$ and an approximate successor features mapping $\psi(\phi_\bs, \ba; \theta_\psi) \approx \psi^{\pi^*}(\phi_{\bs}, \ba)$ (a fully connected network with parameters $\theta_\psi$) by alternating stochastic gradient descent steps on two objective functions:
\begin{align}
&L\underset{(s, a, s') \in \cD_T}{(\theta_\psi) = \ \mathbb{E}} \left[ \left( \phi_\bs + \gamma\psi(\phi_{\bs'}, \ba^*; \theta^-_\psi) - \psi(\phi_{\bs}, \ba; \theta_\psi)\right)^2 \right],
\label{eq:obj_sf}
\end{align}
\begin{equation}
\begin{aligned}
L(&\theta_\phi, \theta_d, \omega) \\
&= \underset{(s, R(s)) \in \cD_R}{ \mathbb{E}} \Big\lbrack \left( R(\bs) - \phi_{\bs}^T \omega \right)^2 + \big( \bs - d(\phi_\bs; \theta_d) \big)^2 \Big\rbrack,
\end{aligned}
\label{eq:obj_phi}
\end{equation}
where $\cD_T$ and $\cD_R$ denote collected experience data for transitions and rewards,
respectively, ${\ba^*} = \argmax_{a'} Q(\bs', \ba'; \pi^*)$ -- computed by inserting the approximate successor features $\psi(\phi_{\bs'}, \ba^*; \theta^-_\psi)$ into Eq. \eqref{eq:srqfac} -- and where $\theta^{-}_\psi$ denotes the parameters of the current target successor feature approximation. To provide stable learning  these are occasionally copied from $\theta_\psi$ (a full discussion of the intricacies of this approach is out of the scope of this paper and we refer to~\cite{mnih2015human} and~\cite{kulkarni2016deep} for details); we replace the target successor feature parameters every $5000$ training steps.

The objective function from Eq. \eqref{eq:obj_sf} corresponds to learning the successor features via online Q-learning (with rewards $\phi(\cdot)$). The objective from Eq. \eqref{eq:obj_phi} corresponds to learning the reward weights and the CNN feature mapping and consists of two parts: the first part ensures that the reward is regressed; the second part ensures that the features are representative of the state $\bs$ by enforcing that an inverse mapping from $\phi(\bs; \theta_\phi)$ to $\bs$ exists through a third convolutional network, a decoder $d(\cdot)$, whose parameters $\theta_d$ are also learned. After learning, actions can be chosen greedily from $Q(\bs, \ba; \pi^*)$ by inserting the approximated successor features into Eq. \eqref{eq:srqfac}.

\section{Transferring successor features to new goals and tasks}
\label{sec:transfer}

\begin{figure}[t]
   \centering
   \includegraphics[width=0.48\textwidth]{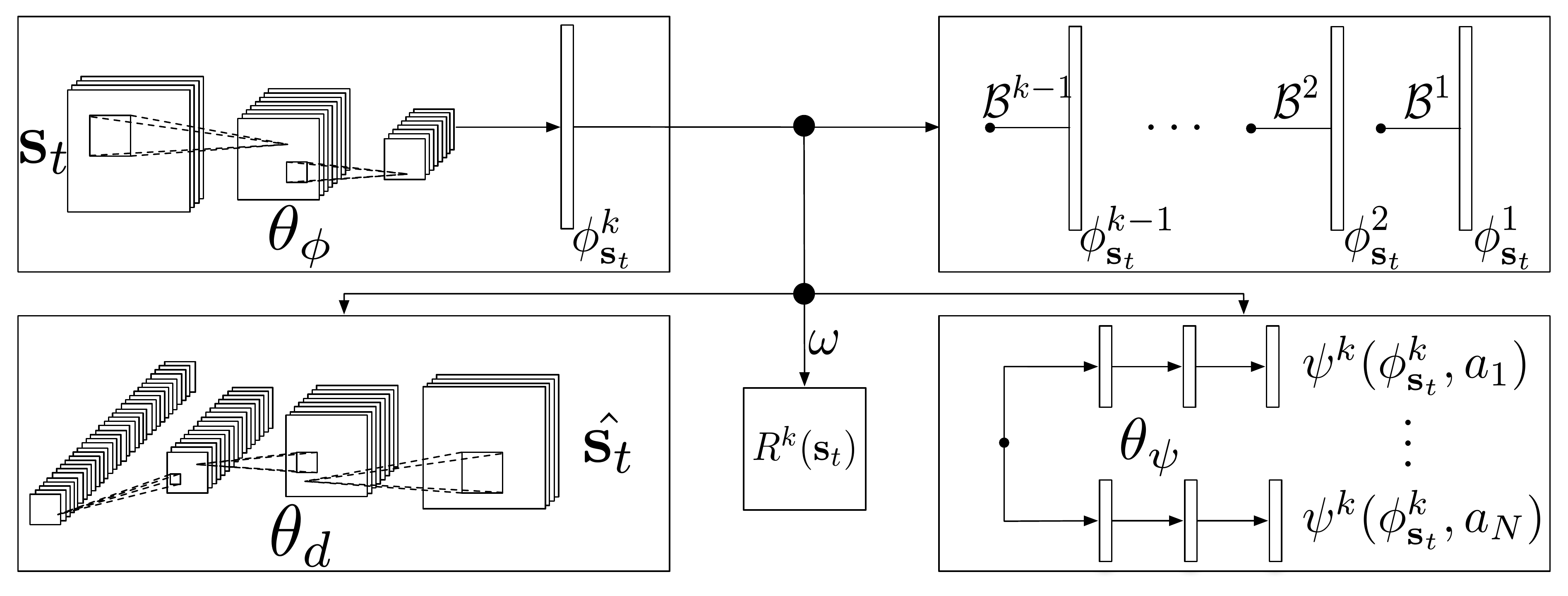}
   \caption{Visualization of the model architecture: $\theta_{\phi}$ parameterizes a convolutional network for extracting features $\phi^k_{\bs_t}$ ($k$ is the current target task) from $\bs_t$ (contains three convolutional layers, with the first layer consisting of 32 $8\times8$ filters with stride 4, the second of 64 $4\times4$ filters with stride 2 and the 3rd of 64 $3\times3$ filters with stride 1, each followed by a rectifying nonlinearity; the last layer is followed by one fully-connected layer with 512 units); $\theta_d$ reconstructs $\bs_t$ back from $\phi^k_{\bs_t}$ (contains five de-convolutional layers, with feature sizes \{512, 256, 128, 64, 4\} and increasing spatial dimensionality in factors of 2); $\omega$ regresses the immediate reward $R^k(\bs_t)$ out of the state representation $\phi^k_{\bs_t}$; $\theta_{\psi}$ computes the successor features $\psi^{k}(\phi^{k}_{\bs_t}, a_n; \theta_{\psi^k})$ for each $a_n \in \cA$ (contains two fully-connected layers); $\cB^i$ maps the features of the current task $k$ back to those of the old tasks.}
   \label{fig:model}
\end{figure}

As described above, the successor representation naturally decouples task specific reward estimation and the estimation of the expected occurrence of the features $\phi(\cdot)$ under the specific policy dynamics. This makes successor feature based RL a natural choice when aiming to transfer knowledge between related tasks. To see this let us first define two notions of knowledge transfer. In both cases we assume that the learning occurs in $K$ different stages during each of which the agent can interact with a distinct task $k \in [1, K]$. The aim for the agent is to solve all tasks at the end of training, using minimal interaction time for each task. From the perspective of reinforcement learning this setup corresponds to a sequence of $K$ RL problems which have shared structure. Knowledge transfer between tasks can then occur for two different scenarios:

The first, and simplest, notion of knowledge transfer occurs if all $K$ tasks use the same environment and transition dynamics and differ only in the reward function $R$. In a navigation task this would be equivalent to finding paths to $K$ different goal positions in one single maze.

The second, and more general, notion of knowledge transfer occurs if all $K$ tasks use different environments (and potentially different reward functions) which share some similarities within their state space. In a navigation task this includes changing the maze structure or robot dynamics between different tasks.

We can observe that successor feature RL lends itself well to transfer learning in scenarios of the first kind: If the features $\phi(\cdot)$ are expressive enough to ensure that the rewards for all tasks can be linearly predicted from them then
for all tasks following the first (i.e., for $k > 1$) one only has to learn a new reward weight vector $\omega^k$ (keeping both the learned $\phi(\cdot)$ and $\psi(\cdot)$ fixed), although care has to be taken if the expectation of the features under the different policy changes (in which case the successor features would have to be adapted also). Learning for $k > 1$ then boils down to solving a simple regression problem (i.e., minimizing Eq.~\eqref{eq:obj_phi} wrt.~$\omega$) and requires only the storage of an additional weight vector per task. This idea has recently been explored ~\cite{barreto2016successor,kulkarni2016deep}. Kulkarni \etal~\cite{kulkarni2016deep} showed large learning speedups for a special case of this setting where they changed the scale of the final reward. 
We here argue that successor feature RL can be easily extended to transfer learning of the second kind with minimal additional memory and computational requirements.

Specifically, to derive a learning algorithm that works for both transfer scenarios let us first define the action-value function for task $k$ using the successor feature notation as
\begin{equation}
    Q^{k}(\bs, \ba; \pi^k) \approx
    \mathbb{E}\left[ \sum_{t=0}^{\infty} \gamma^t \phi^k_{\bs_{t}} \Big| \bs_0=\bs, \ba_0=\ba, \pi^k \right] \cdot \omega^k,
    \notag
    \label{eq:q_tasks}
\end{equation}
where we used the superscript $k$ to refer to task specific features and policies respectively and where we have again introduced the short-hand notation $\phi^k_{\bs_{t}} = \phi^k(\bs_{t}; \theta_{\phi^k})$ for notational brevity.
Additionally, let us assume that there exists a linear relation between the task features, that is there exists a mapping $\phi^{i}_\bs = \cB^i \phi^{k}_\bs$ for all $i \leq k$ and we have $B^k = \bI$. We note that such a linear dependency between features does not imply a linear dependency between the observations (since $\phi(\cdot)$ is a nonlinear function implemented by a neural network), and hence this assumption is not very restrictive. Then -- again using the fact that the expectation is a linear operator --
we obtain for $i \leq k$:
\begin{align}
    Q^{i}(\bs, \ba; \pi^{i}) &\approx
    \mathbb{E}\left[ \sum_{t=0}^{\infty} \cB^i \gamma^t \phi^k_{\bs_{t}} \Big| \bs_0=\bs, \ba_0=\ba, \pi^i \right] \cdot \omega^i \nonumber \\
  &= \cB^i \mathbb{E}\left[ \sum_{t=0}^{\infty} \gamma^t \phi^k_{\bs_{t}} \Big| \bs_0=\bs, \ba_0=\ba, \pi^i \right] \omega^i \nonumber \\
  &= \cB^i \psi^{\pi^i}\left( \phi^k_{\bs_{t}}, \ba \right)^T \omega^i \label{eq:q_tasks_B1} \\
  &= \psi^{\pi^i}\left( \cB^i \phi^k_{\bs_{t}}, \ba \right)^T \omega^i.     \label{eq:q_tasks_B2}
\end{align}
These equivalences now give us a straight-forward way to transfer knowledge to new tasks while keeping the solution found for old tasks intact (as long as we have access to all feature mappings $\phi^k$ and policies $\pi^k$): 
\begin{enumerate}
\item When training on task $k > 1$ initialize the parameters $\theta_{\phi^k}$ and $\theta_{\psi^k}$ with $\theta_{\phi^{k-1}}$ and $\theta_{\psi^{k-1}}$ respectively (otherwise initialize randomly) and train $\psi^{\pi^k}$ and $\phi^k$ via stochastic gradient descent on Eqs.~\eqref{eq:obj_sf}-\eqref{eq:obj_phi}.
\item In addition, train all $\cB^i$ with $i < k$ to preserve the relation $\phi^{i}_\bs \approx \cB^i \phi^{k}_\bs$.
\item To obtain successor features for the previous tasks, estimate the expectation of the features for the current task $k$ under the old task policies to obtain $\psi^{\pi^i}\left( \phi^k_\bs, \ba \right)$ so that Eq.~\eqref{eq:q_tasks_B1} can be computed during evaluation. Note that this means we have to estimate the expectation of the current task features under all old task dynamics and policies\footnote{In principle, the expectations for all tasks $i < k$ need to be evaluated with samples from these tasks. In our case, we however found that the shared structure between tasks was large enough to allow for estimating all expectations based on the current tasks samples only.}. Since we expect significant overlap between tasks in our experiments this can be implemented memory efficiently by using one single neural network with multiple output layers to implement all task specific successor features. Alternatively, if the successor feature networks are small, one can just preserve the old task successor feature networks and use Eq.~\eqref{eq:q_tasks_B2} for selecting actions for old tasks.
\end{enumerate}

When -- as in Sec.~\ref{sec:background_sf} --  we are only interested in finding the optimal policy $\pi^{i^*}$ for each task these steps correspond to alternating stochastic gradient descent steps on two objective functions analogous to Eqs.~\eqref{eq:obj_sf},\eqref{eq:obj_phi}, under the model architecture depicted in Fig.~\ref{fig:model}. More precisely, we write  $\psi^i(\phi^k_\bs, \ba; \theta_{\psi^i}) \approx \psi^{\pi^{i^*}}(\phi^k_{\bs}, \ba)$ and obtain the following objectives for task $k$:

\begin{align}
&\begin{aligned}
&L^k\Big(\lbrace \theta_{\psi^1}, \dots,  \theta_{\psi^{k}} \rbrace \Big) =\\
&\sum_{i \leq k}\underset{\substack{(s, a, s') \\ \in \cD^i_T}}{\mathbb{E}} \left[ \left( \phi^k_\bs + \gamma\psi^i(\phi^k_{\bs'}, \ba^{i*}; \theta^-_{\psi^i}) - \psi^i(\phi^k_{\bs}, \ba; \theta_{\psi^i})\right)^2 \right],
\end{aligned}
\label{eq:obj_sf_k} \\
&L^k\Big(\theta_\phi, \theta_d, \omega^k,  \lbrace \cB^1, \dots, \cB^{k-1} \rbrace \Big) \nonumber =\\
&\begin{aligned}
&\underset{(s, R(s)) \in \cD^k_R}{\mathbb{E}} &\Big\lbrack \left( R(\bs) - {\phi_{\bs}^{k}}^{T} \omega^k \right)^2+ \big( \bs - d(\phi_{\bs}^{k}; \theta_{{d}^{k}}) \big)^2 \Big\rbrack \\
& \ &+ \sum_{i < k} \underset{(s, R(s)) \in \cD^i_R}{\mathbb{E}} \Big\lbrack \big(\phi^i_s - \cB^i \phi^k_s \big)^2  \Big\rbrack,
\end{aligned}
\label{eq:obj_phi_k}
\end{align}
where $a^{i*} = \argmax_{a'} Q(\bs', \ba'; \pi^{i^*})$ is the current greedy best action for task $i$ and in cases where we are willing to store the old $\psi^i(\cdot)$ for $i < k$ Eq.~\eqref{eq:obj_sf_k} only needs to be optimized with respect to $\theta_{\psi^k}$ (dropping all other terms)\footnote{In practice there is no noticeable performance difference.}. Several interesting details can be noted about this formulation. First, if we assume that all $\psi^i(\cdot)$ are implemented using one neural network with $k$ output layers -- or if the successor feature networks are small -- then the overhead for learning $k$ tasks is small (we only have to store $k-1$ additional weight matrices plus one additional reward weight vector per task) this is in contrast to other successful transfer learning approaches for RL that have recently been proposed such as~\cite{rusu2016progressive}. Second, the regression of the old task features via the transformation matrices $\cB^i$ forces the CNN that outputs $\phi^k_s$ to represent the features for all tasks \footnote{May be seen as special case of the distillation technique~\cite{hinton2015distilling}.}. As such we expect this approach to work well when tasks have shared structure; if they have no shared structure one would have to increase the number of parameters (and thus possibly the dimensionality of $\phi^k$).

To gain some intuition for the reasons why the above model should work we here want to give a -- hypothetical -- example:
  Let us assume the set of extracted features $\phi(\cdot)$ to be the relative distance to a set of objects from the current position of the agent. Then, the successor features $\psi(\cdot)$ would estimate the discounted sum of those relative distances under the current policy dynamics. When transferring to a new environment, the spatial relationship of the objects could, for example, change. Then $\phi(\cdot)$ would need to adapt accordingly. But since we assume the two environments to share structure (e.g., they contain the same objects), filters in the early layers of $\phi(\cdot)$ could be largely reused (or transferred). The adapted features (e.g., the relative distances from the current pose to the changed object positions) now would differ from those of the previous environments, this change in scale could be directly captured by a linear mapping $\cB$. $\psi(\cdot)$ would also need to be adapted, but due to the shared structure between environments and their similarity in the successor features we would expect adapting them to be fast. Similarly, the reward mapping can either be re-learned quickly or transferred directly (e.g., if we assume that the reward penalizes proximity to objects).

\section{Simulated Experiments}
\label{sec:experiments}

\subsection{Experimental setup}


\begin{figure}[b]
\begin{minipage}[h]{0.48\linewidth}
\begin{center}
\includegraphics[height=0.49in]{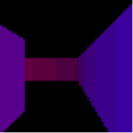}
\vspace{0.1cm}
\includegraphics[height=0.49in]{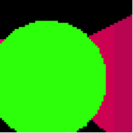}
\vspace{0.1cm}
\includegraphics[height=0.49in]{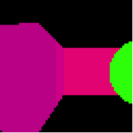}
\end{center}
    \vspace{-.5cm}
\caption{Exemplary views the agent observes in the simulated environment.}
\label{fig:firstperson}
\end{minipage}
\hfill
\begin{minipage}[h]{0.48\linewidth}
\begin{center}
\includegraphics[height=1.in]{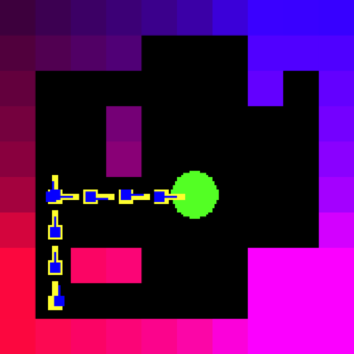}
\end{center}
    \vspace{-.4cm}
\caption{Comparison between the true (yellow) and the predicted (blue) poses.}
\label{fig:pose}
\end{minipage}%
\end{figure}

We first test our algorithm using a simulation of different maze-like 3D environments.
 The environment contains cubic objects and a target for the agent to reach (rendered as a green sphere) (cf.~Fig.~\ref{fig:firstperson}). We model the legal actions as four discrete choices: \{stand still, turn left ($90^\circ$), turn right ($90^\circ$), go straight ($1 m$) \} to simplify the problem (we note that in simulation the agent moves in a continuous manner). The agent is a simulated Pioneer-3dx robot moving under a differential drive model (with Gaussian control noise, thus the robot will have observations of the environment from a continuous viewing position and angle space).

The agent obtains a reward of $-0.04$ for each step it takes, $-0.96$ for colliding with obstacles, $1$ for reaching the goal; this reward structure forces time-optimal behavior. Each episode starts with the agent in a random location and ends when it reaches the goal (unless noted otherwise).

In each time-step the agent receives as an observation a frame captured from the forward facing camera (as shown in Fig.~\ref{fig:firstperson}, re-scaled to $64 \times 64$ pixels). The state in each time-step is then given by the 4 most recently obtained observations.
The top-down views of the four different mazes we consider are shown in Fig.~\ref{fig:noisymaps}.


For training the model (Fig.~\ref{fig:model}) we employed stochastic gradient descent with the ADAM optimizer \cite{Kingma2015}, a minibatch size of $64$ and a learning rate of $2.5\times10^{-4}$ and $2.5\times10^{-5}$ for visual inputs for the supervised learner and the reinforcement learners respectively, $5.0\times10^{-5}$ for depth inputs. We performed a coarse grid search for each learning algorithm to choose the optimizer hyper parameters (learning rate in range $[1\times10^{-6}, 1\times10^{-3}$]) and use the same minibatch size across all considered approaches. Training was performed alongside exploration in the environment (one batch is considered every 4 steps).


\subsection{Baseline method - supervised learning \& DQN }
As a baseline for our experiments, we train a CNN by supervised learning to directly predict the actions computed by an $A^*$ planner from the same visual input that the SF-RL model receives. The network structure is the same as the CNN from the SF-RL model ($\theta_{\phi}$) and differs only in that the output 512 units are fed into a final softmax layer. As an additional baseline we also compare to the DQN approach \cite{mnih2015human}. To ensure a fair comparison we evaluate DQN by learning from scratch and in a transfer learning situation in which we finetune the DQN model trained on the base task; such a fine-tuning approach is known to perform better than simply transferring with fixed features \cite{yosinski2014transferable, rusu2016progressive} (we also conduct transfer learning experiments with fixed features for DQN for completeness).

The training data for the supervised learner is generated beforehand, consisting of $1.6e5$ labeled samples. To generate these samples, full localization is required, while for evaluating the learned network it is not required. As such, this setup can be thought of as the best case scenario for training a CNN to imitate a planner in this domain.

To ensure a fair comparison between different methods in the following plots, we scale the number of steps taken by the supervised learner, so that the number of updates matches that of the SF-RL model and that of DQN (the two reinforcement learners start learning at $3e4$ iterations and makes an update every $4$ steps after that).

\subsection{Visual navigation in 3D mazes}
\begin{figure}[t]
    \centering
    \begin{minipage}{0.48\textwidth}
        \vspace{0.1cm}
        \begin{subfigure}[t]{0.24\textwidth}
            \centering
            \includegraphics[height=.81in]{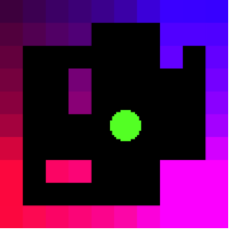}
            \caption{Map1}
            \label{fig:map1}
        \end{subfigure}%
        \vspace{0.1cm}
        \begin{subfigure}[t]{0.24\textwidth}
            \centering
            \includegraphics[height=.81in]{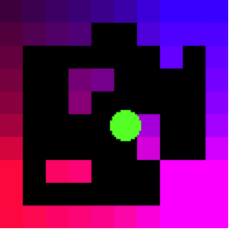}
            \caption{Map2}
            \label{fig:map2}
        \end{subfigure}
        \vspace{0.1cm}
        \begin{subfigure}[t]{0.24\textwidth    }
            \centering
            \includegraphics[height=.81in]{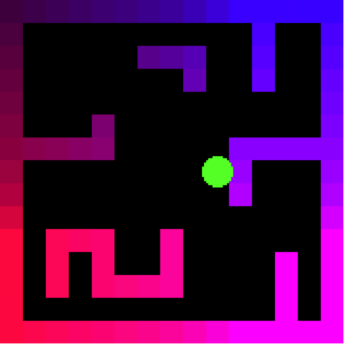}
            \caption{Map3}
            \label{fig:map3}
        \end{subfigure}%
        \vspace{0.1cm}
        \begin{subfigure}[t]{0.24\textwidth}
            \centering
            \includegraphics[height=.81in]{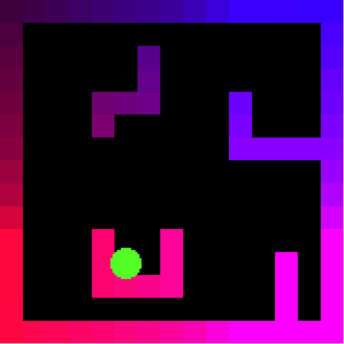}
            \caption{Map4}
            \label{fig:map4}
        \end{subfigure}
        \vspace{0.1cm}
    \end{minipage}
    \vspace{-.5cm}
    \caption{Top-down schematic view of the four different maze environments we consider for the simulated experiments.}
    \label{fig:noisymaps}
    \vspace{-.11cm}
\end{figure}

For the first experiment we trained our deep successor feature reinforcement learner (SF-RL), DQN and the supervised learner on the base map: \textit{Map1} (Fig.~\ref{fig:map1}).
To compare the algorithms we perform a testing phase every 10,000 steps consisting of evaluating the performance of the current policy for 5,000 testing steps.

\subsubsection{Base environment}
We first train on \textit{Map1} from scratch. We observe that the supervised learning and reinforcement learning (DQN and SF-RL) models converge to performance comparable to the optimal $A^*$ planner. We also observe that the supervised learner converges significantly faster in this experiment. This is to be expected since it has access to the optimal paths -- as computed via $A^*$ -- for starting positions covering the whole environment right from the beginning of training. In contrast to this, the reinforcement learners gradually have to build up a dataset of experience and can only make use of the sparsely distributed reward signal to evaluate the actions taken.

\begin{figure}[ht]
    \centering
        \begin{subfigure}[t]{\linewidth}
            \centering
            \includegraphics[width=\linewidth]{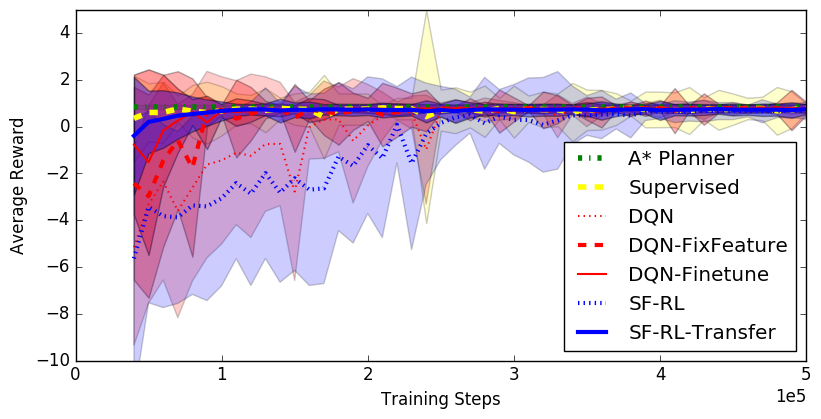}
            \caption{Comparison between learning algorithms for \textit{Map2}.}
            \label{fig:noisyplots}
        \end{subfigure}%

        \begin{subfigure}[t]{\linewidth}
            \centering
            \includegraphics[width=\linewidth]{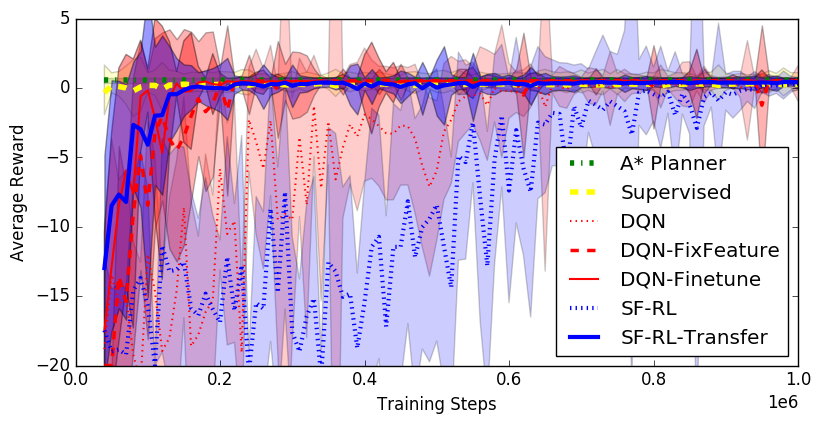}
            \caption{Comparison between learning algorithms for \textit{Map4}.}
            \label{fig:noisyNewplots}
        \end{subfigure}
    \caption{Average reward $\pm$ one standard deviation obtained by $A^*$ using the true system model, the supervised learner, as well as DQN and SF-RL when learning from scratch and with task transfer from \textit{Map1} (\ref{fig:noisyplots}) and \textit{Map3} (\ref{fig:noisyNewplots}).}
    \label{fig:visplots}
    \vspace{-.11cm}
\end{figure}

\subsubsection{Transfer to different environment}
We then perform a transfer learning experiment (using the trained models from above) to a changed environment \textit{Map2} where more walls are added (Fig.~\ref{fig:map2}). In Fig.~\ref{fig:noisyplots} we show a performance comparison between the supervised learner (\textit{Supervised}), DQN learning from scratch (\textit{DQN}) and using task transfer (with fixed CNN layers: \textit{DQN-FixFeature}, and by finetuning the whole network: \textit{DQN-Finetune}), SF-RL from scratch (\textit{SF-RL}) and using task transfer (\textit{SF-RL-Transfer}).
\par
We observe that \textit{SF-RL-Transfer} converges to performance comparable to the optimal policy much faster than training from scratch. Furthermore, in Fig.~\ref{fig:noisyplots} the learning speed of \textit{SF-RL-Transfer} is even comparable to that of \textit{Supervised}, who is learning directly from perfectly labeled actions. We observe that when training from scratch, \textit{DQN} is slightly faster than \textit{SF-RL} (we attribute this to the fact that \textit{SF-RL} optimizes a more complicated loss function including e.g. an auto-encoder loss).
In the transfer learning setting \textit{SF-RL-Transfer} is comparable to \textit{DQN-Finetune}, and converges faster than \textit{DQN-FixFeature}. It is important to realize that our method preserves the ability to solve the old task after this transfer occurred, which \textit{DQN-Finetune} is not capable of. To verify this preservation of the old policies we re-evaluated \textit{DQN-Finetune} and \textit{SF-RL-Transfer} on all tasks and summarize the results in Tab.~\ref{testsing} (\textit{DQN-FixFeature} keeps the network for the initial tasks completely unchanged thus it is unnecessary to evaluate its performance again). We note that our agent is still able to perform well on the old task, while the DQN agent deviated significantly from the optimal policy (it is still able to solve most of the episodes in this case via a ``random-walk'').
\par
We also want to emphasize that in contrast to \textit{DQN-FixFeature}, \textit{SF-RL-Transfer} has the ability of continuously adapting its features to new tasks while keeping a mapping to all previous task features. Additionally, \textit{DQN-FixFeature} has to perform the same transfer procedure for all kinds of transfer scenarios due to its black-box property; while with the flexibility of the more structured representation of \textit{SF-RL-Transfer}, we only need to retrain the successor feature network $\theta_{\psi}$ and keep the reward mapping $\omega$ fixed when only the dynamics changes, or if the dynamics of the environment stay fixed or close to the already observed dynamics, \textit{SF-RL-Transfer} can adapt quickly by either changing only $\omega$ or in combination with $\theta_{\psi}$.

\subsection{More complicated transfer scenarios}

We then experiment in a more complicated transfer scenario: transferring a base controller from \textit{Map3} (Fig.~\ref{fig:map3}) to \textit{Map4} (Fig.~\ref{fig:map4}). As can be seen from the visualization, the objects change significantly from \textit{Map3} to \textit{Map4}. Also, the goal location moves from the center of an open area to a ``hidden'' corner. The results for this experiment are depicted in Fig.~\ref{fig:noisyNewplots}, revealing a similar trend as for the simpler mazes. A re-evaluation of the \textit{DQN-Finetune} and \textit{SF-RL-Transfer} agent is shown in Tab.~\ref{testsing}. We note that the \textit{DQN-Finetune} agent loses the policy for \textit{Map3} after being transferred to \textit{Map4} as the locations of the target and objects changed dramatically, while our agent still is able to solve the old task after the transfer.


Furthermore, when transferring from \textit{Map1} to \textit{Map2} we move from a simpler to a more complicated environment, while \textit{Map4} is ``simpler'' than \textit{Map3}.

\begin{table}[t]
\caption{Final testing statistics for all considered environments, each evaluated from 50 random starting positions. The maximum number of steps per episode was: 200 steps for \textit{Map1\&2}, 500 steps for \textit{Map3\&4}.}
\vspace{-.17cm}
\label{testsing}
\begin{center}
\begin{tabular}{|x{1.52cm}|x{0.9cm}|x{2.1cm}|x{2.1cm}|}
\hline
Pre-train on / Transfer to & Success ratio & Reward & Steps \\
\hline
\end{tabular}
\begin{tabular}{|p{1.52cm}|x{0.9cm}|x{2.1cm}|x{2.1cm}|}
\multicolumn{4}{c}{Testing on Map1}\\
\hline
$A^*$ baseline & & 0.814 $\pm$ 0.070 & 5.640 $\pm$ 1.747 \\
\hline
\multicolumn{4}{|c|}{DQN-Finetune}\\
\hline
Map1 / -    & 50/50 & 0.791 $\pm$ 0.114 & 6.220 $\pm$ 2.845 \\ 
Map1 / Map2 & 48/50 & 0.398 $\pm$ 1.755 & 15.000 $\pm$ 38.800 \\
\hline
\multicolumn{4}{|c|}{SF-RL-Transfer}\\
\hline
Map1 / -    & 50/50 & 0.765 $\pm$ 0.243 & 6.410 $\pm$ 3.915 \\ 
Map1 / Map2 & 50/50 & 0.733 $\pm$ 0.235 & 6.796 $\pm$ 2.999 \\
\hline
\end{tabular}
\begin{tabular}{|p{1.52cm}|x{0.9cm}|x{2.1cm}|x{2.1cm}|}
\multicolumn{4}{c}{Testing on Map3}\\
\hline
$A^*$ baseline & & 0.635 $\pm$ 0.138 & 10.120 $\pm$ 3.438 \\
\hline
\multicolumn{4}{|c|}{DQN-Finetune}\\
\hline
Map3 / -    & 50/50 & 0.566 $\pm$ 0.178 & 11.84 $\pm$ 4.442 \\
Map3 / Map4 &  4/50 & -18.335 $\pm$ 5.703 & 460.46 $\pm$ 135.450 \\
\hline
\multicolumn{4}{|c|}{SF-RL-Transfer}\\
\hline
Map3 / -    & 50/50 & 0.489 $\pm$ 0.348 & 13.460 $\pm$ 5.936 \\
Map3 / Map4 & 50/50 & 0.444 $\pm$ 0.416 & 13.780 $\pm$ 8.707 \\
\hline
\end{tabular}
\end{center}
    \vspace{-.11cm}
\end{table}

\subsection{Analysis of learned representation}
As an additional test, we analyzed the representation $\phi^k_s$
learned by the SF-RL approach. Specifically, since the reward is
defined on the pose of the agent and optimal path finding clearly
depends on the agent being able to localize itself we analyzed as to
whether $\phi^k_s$ encodes the robot pose. To answer this, we
extract features $\phi^k_s$ for all
states along collected optimal trajectories and regressed the
ground truth poses of the robot (obtained from our simulator)
using a neural network with two hidden
layers (128 units each). Fig.~\ref{fig:pose} shows the results
from this experiment, overlaying the ground truth poses with the
predicted poses from our regressor on a held out example. From these
we can conclude that indeed, the transition dynamics is encoded and
the agent is able to localize itself, and this information
can reliably be retrieved post-hoc (i.e., after training).

\section{Real-World Experiments}
\label{sec:}

In order to show the applicability of our method to more realistic
scenarios, we conducted additional experiments using a real robot. We
start by swapping the RGB camera input for a simulated depth sensor in
simulation and then perform a transfer learning experiment to a
different, real, environment from which we collect real depth images.

\subsection{Rendered Depth Experiments}
\begin{figure}
    \begin{subfigure}{0.48\textwidth}
        \centering
        \includegraphics[height=1.58in]{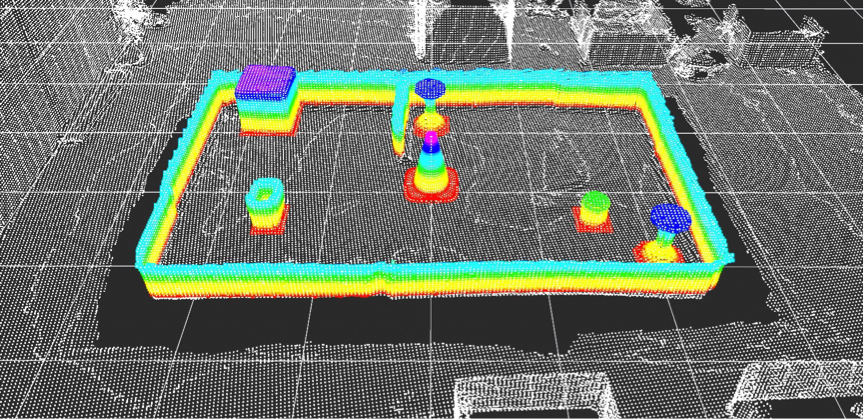}
    \end{subfigure}
    \vspace{0.1cm}
    \hrule
    \vspace{0.1cm}
    \begin{subfigure}{0.48\textwidth}
        \includegraphics[height=0.38in]{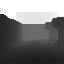}
        \includegraphics[height=0.38in]{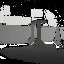}
        \includegraphics[height=0.38in]{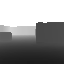}
        \includegraphics[height=0.38in]{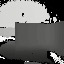}
        \includegraphics[height=0.38in]{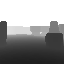}
        \includegraphics[height=0.38in]{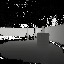}
        \includegraphics[height=0.38in]{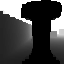}
        \includegraphics[height=0.38in]{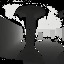}
    \end{subfigure}
    \caption{3D Model of \textit{Map5}. The robot should avoid the
      colored regions containing objects and navigate to the traffic
      cone in the center. The bottom part shows pairs of images comparing
      between the rendered depth images from our simulator, and real
      depth images taken by a kinect camera at approximately the same
      pose in a real environment modeled
      after the simulator.}
    \label{fig:rendered-depth}
    \vspace{-.13cm}
\end{figure}

To obtain a scenario more similar to a real world scene we
might encounter, we build a maze-like environment \textit{Map5}
(Fig. \ref{fig:rendered-depth}) in our robot simulator that includes
realistic walls and object models. In this setting the robot has to
navigate to the target (traffic cone in the center) and avoid colliding with
objects and walls. We then simulate the robot within this environment,
providing rendered depth images from a simulated kinect camera as the input modality (as opposed to the artificial RGB images we used before).

\subsection{Real World Transfer Experiments}
We then change to a real robot experiment in which the robot can
explore the maze depicted in \textit{Map6} (Fig. \ref{fig:title-real})
(note that the position of the objects and the target are changed from
the simulated environment \textit{Map5} in Fig. \ref{fig:rendered-depth}). We collect real depth images in the
actual maze-world using the on-board kinect sensor of a Robotino. To avoid training
for long periods of time in the real environment we pre-recorded
images at all locations that the robot can explore (taking 100
images per position and direction with randomly perturbed robot pose to model
noise).

\begin{figure}[t]
    \centering
    \includegraphics[width=0.45\textwidth]{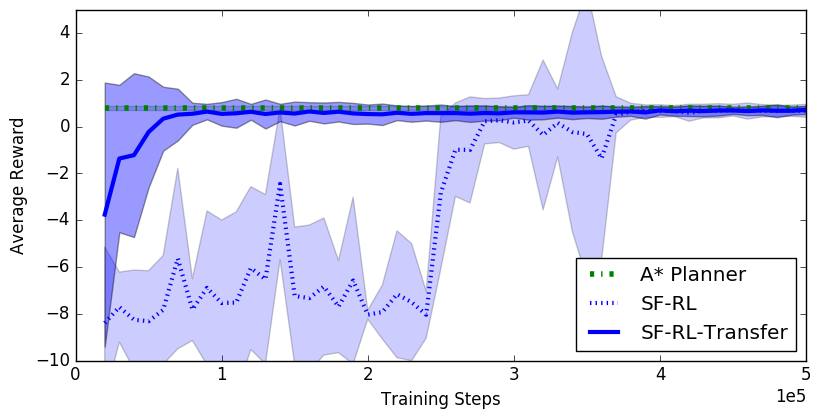}
    \caption{Comparison between SF-RL trained on the real world \textit{Map6}, and SF-RL-Transfer with the base model trained on the simulated \textit{Map5} transferred to \textit{Map6}.}
  \label{fig:realplots}
    \vspace{-.13cm}
\end{figure}

The results of training from scratch in this real environment as well
as when transfer from the simulated environment is performed are
depicted in Fig. \ref{fig:realplots} (the agent starts to learn here after $1e4$ steps, whereas in previous experiments this number is set to $3e4$). Similar to the previous
experiments we see a large speed-up when transferring knowledge even
though the simulated depth images contain none of the characteristic
noise patterns present in the real-world kinect data. We note that
  the agent achieves satisfactory performance at around 60,000
  iterations, which corresponds to approximately 8 hours of real
  experience (assuming data is collected at a rate of 2Hz). 

After training with the pre-recorded images, the robot is tested
in real world environments.
A video of the real experiments in two changed environments: \textit{Map6} \& \textit{Map7} (\textit{Map7} is not discussed here due to space constraints) can be found at: \url{https://youtu.be/WcCcdkhgjdY}.

\section{Conclusion}
\label{sec:conclusionAndFutureWorks}
We presented a method for solving robot navigation tasks from raw sensory data, based on an extension of the theory behind successor feature reinforcement learning. Our algorithm is able to naturally transfer knowledge between related tasks and yields substantial speedups over deep reinforcement learning from scratch in the experiments we performed. Despite of these encouraging results, there are several opportunities for future work including testing our approach in more complicated scenarios and extending it to more naturally handle partial observability.







\small
\bibliographystyle{IEEEtran}
\bibliography{zhang17icra}

\end{document}